# Lexicalization and Grammar Development

B. Srinivas, D. Egedi, C. Doran and T. Becker

Institute for Research in Cognitive Science
University of Pennsylvania, Philadelphia, PA 19104-6228, USA
{srini, egedi, cdoran, tilman}@unagi.cis.upenn.edu

**Abstract.** In this paper we present a fully lexicalized grammar formalism as a particularly attractive framework for the specification of natural language grammars. We discuss in detail Feature-based, Lexicalized Tree Adjoining Grammars (FB-LTAGs), a representative of the class of lexicalized grammars. We illustrate the advantages of lexicalized grammars in various contexts of natural language processing, ranging from wide-coverage grammar development to parsing and machine translation. We also present a method for compact and efficient representation of lexicalized trees.

In diesem Beitrag präsentieren wir einen völlig lexikalisierten Grammatikformalismus als eine besonders geeignete Basis für die Spezifikation von Grammatiken für natürliche Sprachen. Wir stellen feature-basierte, lexikalisierte Tree Adjoining Grammars (FB-LTAGs) vor, ein Vertreter der Klasse der lexikalisierten Grammatiken. Wir führen die Vorteile von lexikalisierten Grammatiken in verschiedenen Bereichen der maschinellen Sprachverarbeitung aus; von der Entwicklung von Grammatiken für weite Sprachbereiche über Parsing bis zu maschineller Übersetzung. Wir stellen außerdem eine Methode zur kompakten und effizienten Repräsentation von lexikalisierten Bäumen vor.

## 1 Introduction

Lexicalized grammar formalisms are particularly well-suited to the specification of natural language grammars. While the use of lexicalization in formal theories is not new, in recent linguistic formalisms the lexicon has played an increasingly important role (e.g. LFG [12], GPSG [7], HPSG [14], CCG [19], Lexicon-Grammars [8], LTAG [16], Link Grammars [18], and some versions of GB [5]). Of these, only CCG, LTAG and Link Grammars are fully lexicalized (this notion will be defined more formally in Section 2).

In this paper, we will be focusing on grammar development and processing aspects of Feature-based, Lexicalized Tree Adjoining Grammars (FB-LTAGs) and the benefits of lexicalization. We will begin with a brief introduction to lexicalized grammars in general and then describe the FB-LTAG formalism in particular. We will then illustrate the advantages of a lexicalized grammar for grammar development in the context of an instantiation of the FB-LTAG formalism for English. One disadvantage of lexicalized grammars is that they are made up of many more elementary structures than their non-lexicalized counterparts and FB-LTAG is no exception. In Section 5 we describe a method for representing FB-LTAGs compactly. In Section 6 we illustrate how lexicalization can be taken advantage of in the parsing of FB-LTAGs. Section 7 describes the use of FB-LTAG in machine translation.

## 2  Lexicalized Grammars

Lexicalized grammars systematically associate each elementary structure with a lexical anchor. This means that in each structure at least one lexical item is realized. The resulting elementary structures specify larger domains of locality over which constraints can be stated (as compared to CFGs).

Following [15] we say that a grammar is **lexicalized** if it consists of:

1. A finite set of structures each associated with a lexical item.
2. An operation or operations for combining the structures.

Each lexical item is called the **anchor** of the corresponding structure over which it specifies linguistic constraints. Hence, the constraints are local to the anchored structure.

## 3  The Feature-Based Lexicalized Tree Adjoining Grammar (FB-LTAG) Formalism

Feature-Based Lexicalized Tree Adjoining Grammars (FB-LTAGs) trace their lineage to Tree Adjunct Grammars (TAGs), which were first developed in [10] and later extended to include unification-based feature structures [20, 21] and lexicalization [15]. Tree Adjoining Languages (TALs) fall into the class of mildly context-sensitive languages, and as such are strictly more powerful than context-free languages. The TAG formalism in general, and lexicalized TAGs in particular, are well-suited for linguistic applications. As first shown by [9] and [13], the properties of TAGs permit us to encapsulate diverse syntactic phenomena in a very natural way. For example, TAG's extended domain of locality and its factoring of recursion from local dependencies lead to a localization of so-called unbounded dependencies.

The primitive elements of the standard TAG formalism are known as **elementary trees**. Elementary trees are of two types: **initial trees** and **auxiliary trees**. In describing natural language, **initial trees** are minimal linguistic structures that contain no recursion, i.e. trees containing the phrasal structure of simple sentences, NPs, PPs, and so forth. Initial trees are characterized by the following: 1) all internal nodes are labeled by non-terminals; 2) all leaf nodes are labeled by terminals or by non-terminal nodes marked for substitution (↓). Recursive structures are represented by **auxiliary trees**, which represent constituents that are adjuncts to basic structures (e.g. adverbials). Auxiliary trees are characterized as follows: 1) all internal nodes are labeled by non-terminals; 2) all leaf nodes are labeled by terminals or by non-terminal nodes marked for substitution, except for exactly one non-terminal node, which is marked as the foot node (∗); 3) the foot node has the same label as the root node of the tree.

Each internal node of an elementary tree is associated with two feature structures (FS), the top and the bottom. The bottom FS contains information relating to the subtree rooted at the node, and the top FS contains information relating to the supertree at that node. Substitution nodes have only a top FS, while all other nodes have both a top and bottom FS.

There are two operations defined in the FB-LTAG formalism, **substitution** and **adjunction**.[1] In the **substitution** operation, a node marked for substitution in an elementary tree is replaced by another elementary tree whose root label is the same as the non-terminal. The features of the node at the substitution site are the unified features of the original nodes. The top FS of the node is the result of unification of the top features of the two original nodes, while the bottom FS of the new node is simply the bottom features of the root node of the substituting tree (since the substitution node has no bottom feature). Figure 1(a) shows two elementary trees and the tree resulting from the substitution of one tree into the other.

---

[1] Technically, substitution is a specialized version of adjunction, but it is useful to make a distinction between the two.

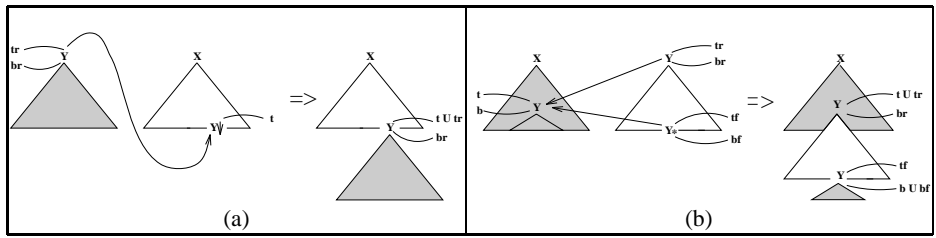

**Fig. 1.** Substitution and Adjunction in FB-LTAG

In an **adjunction** operation, an auxiliary tree is inserted into an elementary tree. The root and foot nodes of the auxiliary tree must match the node label at which the auxiliary tree adjoins. The node being adjoined to splits, and its top FS unifies with the top FS of the root node of the auxiliary tree, while its bottom FS unifies with the bottom FS of the foot node of the auxiliary tree. Figure 1(b) shows an auxiliary tree and an elementary tree, and the tree resulting from an adjunction operation.

The XTAG System [6] is based on FB-LTAG.[2] It contains a wide-coverage English grammar which has been used to parse the Wall Street Journal, IBM manual, and ATIS corpora, with a average coverage of 60%.

## 4 Specifying a Lexicalized Grammar

The specification of a grammar in FB-LTAG differs from that of non-lexicalized grammar formalisms. The syntactic information in a lexicalized grammar is contained in the lexical items themselves, rather than in outside rules that are combined first with each other with the lexical items added in later. This difference in conceptualization of the grammar allows grammar writers to work in a bottom-up style - starting with the words themselves and how they interact. We believe that this is a more natural way of viewing natural language grammars, and it allows grammar developers to easily specify analyses for various phenomena that might otherwise require a more complex analysis. In addition, the use of both feature structures and lexicalization allows the grammar

---

[2] XTAG was developed at the University of Pennsylvania, and consists of: an Earley-style parser, a trigram part-of-speech tagger, a morphological database containing approximately 317,000 inflected items, a syntactic lexicon containing over 37,000 entries, and a tree database with approximately 385 trees composed into 40 tree families which was developed using the XTAG tools. The XTAG grammar covers a wide range of phenomena that includes auxiliaries (both inverted and non-inverted), copula, raising and small clause constructions, topicalization, relative clauses, infinitives, gerunds, passives, adjuncts, it-clefts, wh-clefts, PRO constructions, noun-noun modifications, extraposition, determiner phrases, genitives, negation, determiner ordering, noun-verb contractions and imperatives. The XTAG workbench provides a graphical user interface based on the X11 Windows system and is implemented in Common Lisp. The morphology, grammar and workbench are freely available by anonymous ftp; for more information, send email to xtag-request@linc.cis.upenn.edu.

to be specified more succinctly, without loss of readability, than might otherwise be possible.

The XTAG grammar consists of two components - a tree database and a syntactic lexicon. The separation of these two parts of the grammar allows for a more efficient representation in the syntactic lexicon. Entries for lexical items map directly to trees in the tree database, which contains all of the trees available to the lexical items. The trees naturally fall into two conceptual classes. The smaller class consists of individual trees, as seen in Figures 2(a), 2(d), and 2(e). These trees are generally anchored by lexical items other than main verbs. The larger class consists of trees that are grouped into **tree families**. These tree families represent subcategorization frames; the trees in a tree family would be related to each other transformationally in a movement-based approach. The trees in Figures 2(b) and 2(c) are members of two distinct tree families. Tree 2(b) is in the transitive tree family, while Tree 2(c) is in the sentential complement tree family. Trees 2(d) and 2(e) illustrate the feature structures that are associated with trees.

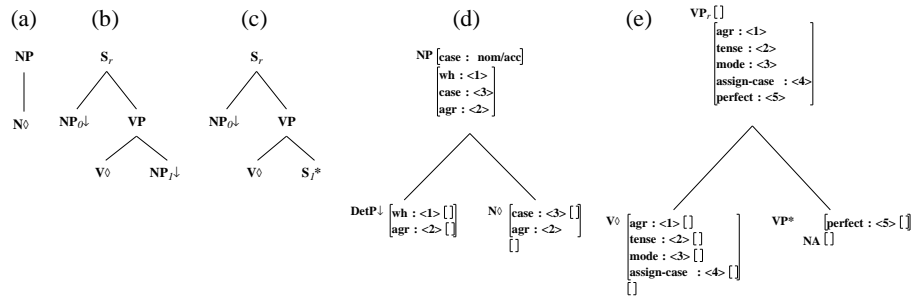

**Fig. 2.** Sample trees from the Tree Database

The syntactic lexicon is one of the most important elements of a lexicalized grammar system. It contains all of the syntactic information about a lexical item, including which trees it selects and what constraints it places on these trees. The use of feature structures in lexicalized grammars provides an efficient way to specify lexical idiosyncrasies for the various lexical items without an explosion of trees. Feature structures allow generalizations across lexical types to be factored out of the lexicon and captured in the trees themselves.

As an example, consider the verb *think*. In addition to being both a transitive and intransitive verb, it can also take an infinitive, indicative, or predicative sentential complement. All three of these possibilities can be handled within the same tree family with different constraints imposed on the sentential complement, according to its feature structures. Table 1 shows syntactic entries for *think* that require a sentential complement. Each of the entries in the FS specifies a constraint that is placed on the $S_1$ node of the sentential complement tree shown in Figure 2(c).

| | | | |
|---|---|---|---|
| INDEX: | think | INDEX: | think |
| ENTRY: | think | ENTRY: | think |
| POS: | Verb | POS: | Verb |
| FRAME: | Sentential_Complement | FRAME: | Sentential_Complement |
| FS: | Indicative_Complement | FS: | Infinitive_Complement |
| EX: | Max thought that the paper was done. | EX: | Doug thought to clean the kitchen. |

**Table 1.** Selected Syntactic Database Entries

As noted above, features are useful for specifying a grammar concisely. One of the initial concerns about FB-LTAGs was that an unlimited number of non-linguistically motivated features would be needed to specify a large grammar. In developing the XTAG system, we have not found this to be true. XTAG uses 30 features to handle a large number of phenomena (listed in footnote 2), and of these only two[3] are not purely linguistically motivated. Some example features are **agr**, **assign-case**, **case**, **conditional, definite, neg, passive, pron, tense,** and **wh**.

In addition, lexicalized grammars facilitate the difficult task of correctly specifying the lexical order of various parts of speech that exhibit 'stacking', that is, lexical items that stack on top of each other in some regular order. Auxiliaries, determiners and adjectives all exhibit this phenomena, as exemplified in the following sentences.

(1) *John should have been waiting.*
(2) *Only these few men agreed to return.*
(3) *The tiny old shabby house finally fell apart.*

The fact that each lexical item carries with it information about which trees it selects, as well as what constraints can be placed on various nodes of those trees, makes specifying the order of the trees straight-forward.

In sequences of auxiliary verbs, for example, each auxiliary verb imposes restrictions on the type of verb that can succeed it. In English, sequences of up to five verbs are allowed, as in the sentence:

(4) *The music should have been being played [when the president arrived].*

The required ordering of these verb forms is:

**modal base past_participle gerund past_participle**

Figure 3 shows the trees anchored by each of the verbs in the sentence.[4] Note that each verb "knows" what its **mode** is, as well as what **mode** its successor must be; this guarantees that only the correct order of auxiliaries is licensed by the grammar.

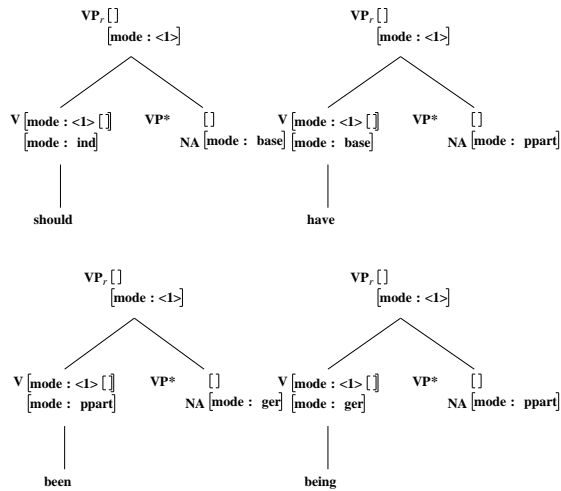

**Fig. 3.** Auxiliary trees for *The music should have been being played.*

The analysis of idioms is another phenomena for which lexicalized grammars are naturally suited. One need only allow elementary structures to have multiple anchors, which are the frozen words of the idiom. This allows for the flexible specification of degrees of idiomaticity, i.e. all leaf nodes are filled for completely frozen expressions, while leaf nodes of expressions which retain some amount of variability can be left open, allowing for limited derivational composition. See [3] for a more detailed discussion.

---

[3] One is used to simulate multi-component adjunction, the other passes inversion information.
[4] For readability, only the relevant features are shown.

## 5 Compact Representation of the Lexicon

An apparent disadvantage of a lexicalized grammar is the proliferation of trees. The XTAG grammar for English contains about 350 different trees which share a large number of similarities. This section sketches metarules which can be used for automatic generation of the trees within each tree family.

There are two well-known mechanisms for a compact representation of grammars: an inheritance network and lexical rules. Both are suitable for lexicalized grammars and have been developed for a number of lexicalized formalisms. They not only permit a compact representation but also capture linguistic generalizations.

In an inheritance network, the syntactic information associated with a particular type of lexical entry can be represented as a combination of the syntactic information for other types with additional information. Proposals for the use of inheritance networks in a TAG grammar can be found in [22, 4].

Lexical rules, called **metarules** in the XTAG framework, are used to capture the similarities between different trees for a lexical entry which cannot be made explicit in the basic framework. They can cover morphological as well as syntactic phenomena. As an example, the relation between an indicative sentence and a clause with wh-movement can be expressed by one metarule for all verbs.

### 5.1 Metarules

In the XTAG grammar the trees for a class of verbs (which have the same subcategorization frame) are grouped in a **tree family**. Tree families include variations such as wh-questions, relative clauses, topicalized, and passive sentences.

An example of a possible metarule in the XTAG English grammar will serve to illustrate the concept of metarules. Almost all tree-families in the grammar include a tree for a wh-question on the subject of the sentence. Figure 4 shows the basic tree (1a) and the wh-tree (1b) for the transitive tree family, which is selected by verbs such as *eat*. Tree (1a) can be used to generate a sentence such as *John eats the cake*, while tree (1b) can be used to generate *Who eats the cake?*

Another tree family which includes a tree for wh-questions (on the subject) is the intransitive tree family, which is selected by verbs such as *sleep*. Figure 4 shows the corresponding basic and wh-trees (2a) and (2b) for this tree family.

Although the trees (1b) and (2b) are not identical, they share the same differences from the corresponding basic tree in their tree family. Both have an additional S-node that is the root and both have a left branch that has an NP-node with the feature **[wh: +]**. This generalization holds for almost all other tree families in the XTAG English grammar and can be expressed as a metarule for wh-movement of the subject. Without giving a formal definition in this paper (see [4]), such a metarule is illustrated in Figure 4.

By using metarules in this style, the number of trees that have to be stated in an FB-LTAG can be reduced considerably. Ideally, for every tree family only one representative tree, which could be called the **base** tree, has to be given; all the other trees can be derived by the application of metarules.

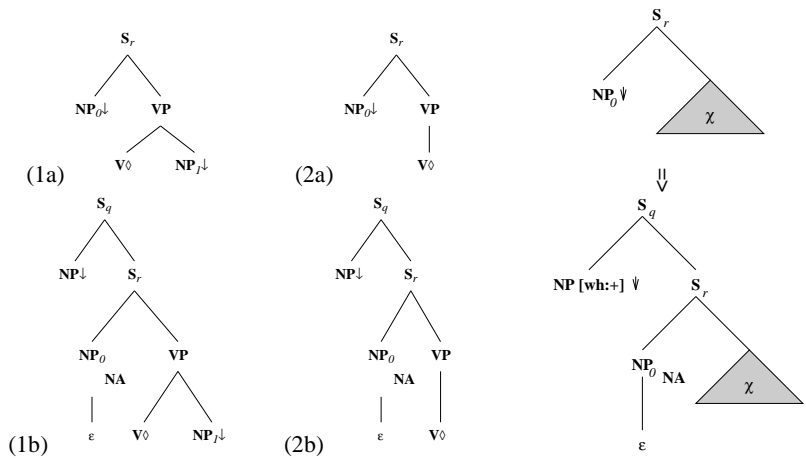

**Fig. 4.** Two pairs of trees from the transitive and the intransitive tree family and a metarule for wh-movement of the subject.

### 5.2 Generative Power of Metarules

In other formalisms which have been enhanced with metarules or similar devices, the resulting framework generally turned out to be undecidable. This results from recursively applicable metarules which can create an infinite number of rules (or trees in a TAG).[5]

In the XTAG framework there is an clear restriction on the application of metarules which derives from the basic principles of TAGs: (i) TAGs factor out recursion with auxiliary trees and (ii) an initial tree describes the extended domain of locality of its anchor.

For the use of metarules, (i) implies that unbounded recursive phenomena should not be described with metarules but rather with auxiliary trees and adjoining and (ii) implies that the output of a metarule should not describe structures beyond the domain of locality of the anchor. Restriction (ii) can be implemented as a predicate which rejects trees that go beyond the domain of locality, thereby allowing a finite number of (recursive) applications but preventing an unbounded application of metarules.

## 6 Tree Disambiguation Using Local Information

One concern in parsing with lexicalized grammars is that the number of elementary structures to be considered by the parser can be quite large. In the worst case it can include all the structures anchored by all the words of the string. In FB-LTAG, however, the richness of the elementary structures allows efficient filtering mechanisms to be used on the set of selected trees, thus passing the fewest possible trees to the parser itself.

As described above, each lexical item in FB-LTAG is associated with one or more elementary trees (one for each syntactic environment in which the item occurs). We call these elementary trees **supertags**, in order to distinguish them from the less complex usual parts of speech. Given an input sentence, the FB-LTAG parser [15] selects the set of supertags for each lexical item of the input sentence and attempts to combine the supertags by searching a large space of supertag combinations to finally yield the parse of the sentence. Eventually, when the parse is complete, there is only one supertag for each word (assuming no global ambiguity).

---

[5] One way around this problem is the introduction of a 'finite closure' restriction, as introduced in GPSG [7]. The underlying assumption is that no metarule can (or needs) to be applied more than once in the process of deriving grammar rules by repeatedly applying metarules to a core grammar rule. Such an arbitrary restriction is problematic, e.g. for languages with multiple wh-movement.

Alternatively, rather than forcing the parser to handle this disambiguation task, we can eliminate (or substantially reduce) the supertag assignment ambiguity prior to parsing by using local information such as local lexical dependencies. As in standard part-of-speech disambiguation, we can use local statistical information, such as bigram and trigram models based on the distribution of supertags in a FB-LTAG parsed corpus. Since the supertags encode dependency information, we can also use information about the distribution of distances of the dependent supertags for a given supertag.

### 6.1 Example of Supertag Disambiguation

The following example illustrates the process of supertag disambiguation for the sentence *John saw a man through the telescope*. Figure 5 lists a few members of the supertag set[6] for each lexical item in the sentence. Figure 5 also shows the final supertag sequence assigned by the supertagger, which picks the best supertag sequence using statistical information about individual supertags and their dependencies on other supertags. The chosen supertags are then combined by substitution and adjunction to obtain a parse.

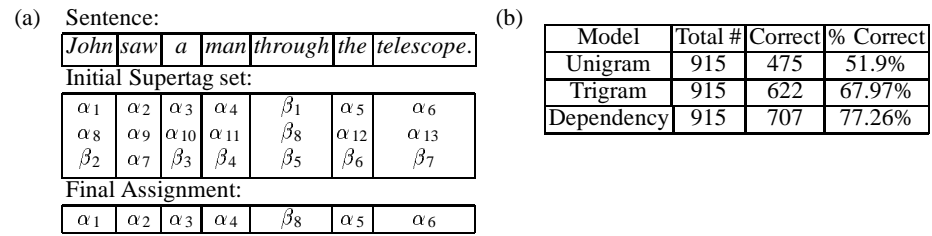

**Fig. 5.** Supertagging and Performance Results of various models of Supertagging

### 6.2 Models, Experiments and Results

We have experimented with unigram and trigram, as well as dependency models for supertag disambiguation. In an n-gram model, dependencies between the supertags that appear beyond the fixed n-word window cannot be incorporated. This limitation can be overcome if no a priori bound is set on the size of the window, and instead a probability distribution of the distances of the dependent supertags[7] for each supertag is maintained. This is the basis for the dependency model. The performance results of the n-gram and dependency models as tested on 100 Wall Street Journal sentences are shown in Figure 5(b). A supertag assignment by a model is considered correct if the word is assigned the same supertag as it would be in the parse of the sentence in which the word appears. The training data for these experiments were collected from the derivation trees of Wall Street Journal sentences that were parsed using XTAG. A detailed presentation of these models and experiments can be found in [11].

---

[6] Each $\alpha$ represents an initial tree and each $\beta$ represents an auxiliary tree.

[7] A supertag is **dependent** on another supertag if the former substitutes or adjoins into the latter.

## 7 Machine Translation using Synchronous FB-LTAGs

Synchronous FB-LTAGs are a variant of FB-LTAGs introduced by [17] to characterize correspondences between Tree Adjoining Languages (TALs). They can be used in either transfer or interlingua-based machine translation by relating TAGs for two different languages [1], or by relating a syntactic TAG and a semantic one for the same language [17, 2]. Although they can be used with either approach, they lend themselves more easily to transfer-based machine translation.

Machine translation is a promising application area for FB-LTAGs. The wider domain associated with each lexical item in FB-LTAGs provides an advantage over exclusively lexical transfer-based approaches in that structures associated with each lexical item can be translated along with the words themselves. The advantages of factoring out recursion and dependencies from the grammar in FB-LTAGs hold in the synchronous FB-LTAG formalism as well. In addition to having individual elementary trees in which these properties hold, we now have pairs of trees, each of which satisfies these properties. The structure of FB-LTAGs allows syntactic dependencies, such as agreement, subcategorization, and so forth, to be localized in the lexical items and their associated elementary trees. This is also true for long-distance dependencies such as wh-phrases. Since transfer rules are stated as correspondences between nodes of the elementary trees associated with lexical entries, we can define lexical transfer rules over a larger domain of locality.

The correspondences between different languages take place on several levels in the synchronous FB-LTAG formalism.

**Elementary Tree Nodes** These correspondences, which are defined for each elementary tree pair, completely specify the target derivation tree given a source derivation tree.

**Lexical Entries** The lexical correspondence is defined in terms of the lexical entries. This is the core of the any transfer-based approach.

**Features** This correspondence includes any morphological features, as well as any other feature restrictions placed on the trees by the lexical item.

## 8 Conclusions

The importance of the lexicon in language processing has been well established. Lexicalized grammars attempt to make the lexicon the central component of the grammar by including in the grammar precisely those structures that are licensed by lexical items in the lexicon. Thus every structure in a fully lexicalized grammar is associated with a lexical item and the lexical item imposes syntactic and semantic constraints on the associated structure. This paper illustrates the advantages of using a lexicalized grammar, such as FB-LTAG, in tasks such as wide-coverage grammar development, parsing and machine translation.